%% file: rome_applied.tex
\IfFileExists{applied_version/brand.sty}{\def\PaperRoot{applied_version/}}{\def\PaperRoot{}}
\PassOptionsToPackage{dvipsnames,table}{xcolor}

\makeatletter
\ifx\PaperRoot\@empty\else
\@for\PaperFile:=acmart.cls,appliedpaper.sty,brand.sty,header.sty,docspec.sty\do{%
  \edef\PaperRedirect{\noexpand\declare@file@substitution{\PaperFile}{\PaperRoot\PaperFile}}%
  \PaperRedirect}
\fi
\@for\PaperFile:=natbib.sty,fancyhdr.sty\do{%
  \edef\PaperRedirect{\noexpand\declare@file@substitution{\PaperFile}{\PaperRoot vendor/\PaperFile}}%
  \PaperRedirect}
\makeatother
\documentclass[acmsmall,nonacm,screen,11pt]{acmart}
\acmYear{2026}
\usepackage[layout=paper]{appliedpaper}
\input{\PaperRoot preamble.tex}

\setpaperteaser{\input{\PaperRoot figures/fig_teaser}}

\begin{document}
\title{Building Rome from a Single Image}
\author{Jiraphon Yenphraphai}
\author{Fang Li}
\author{Tianshuo Xu}
\author{Depu Meng}
\author{Quentin Herau}
\author{Yihan Hu}
\author{Raymond A. Yeh}
\author{Wei Zhan}
\compactauthors{%
  {\raggedright
   {\rmfamily\bfseries
    \appliedname{Jiraphon Yenphraphai\textsuperscript{1, 2*}} \;
    \appliedname{Fang Li\textsuperscript{1, 3*}} \;
    \appliedname{Tianshuo Xu\textsuperscript{1}} \;
    \appliedname{Depu Meng\textsuperscript{1}} \\[0.25em]
    \appliedname{Quentin Herau\textsuperscript{1}} \;
    \appliedname{Yihan Hu\textsuperscript{1}} \;
    \appliedname{Raymond A. Yeh\textsuperscript{2}} \;
    \appliedname{Wei Zhan\textsuperscript{1, 4\dag}}\par}
   \smallskip
   {\small\rmfamily\color{AppliedGray}%
    \textsuperscript{1}Applied Intuition \quad
    \textsuperscript{2}Purdue University\\
    \textsuperscript{3}University of Illinois at Urbana-Champaign \quad
    \textsuperscript{4}UC Berkeley\par}
   \smallskip
   {\footnotesize\rmfamily\color{AppliedGray}%
    \textsuperscript{*}Work done during an internship at Applied Intuition \quad
    \textsuperscript{\dag}Corresponding author: \href{mailto:wei.zhan@applied.co}{wei.zhan@applied.co}\par}

    \medskip
   {\centering
    \small
    \begin{center}
{\url{https://build-rome.github.io/}}
\end{center}\par}
    \vspace{-0.6cm}
   \par}
}

\begin{abstract}
\input{\PaperRoot src/abstract}
\end{abstract}
\makefrontmatter
\input{\PaperRoot src/intro_new}
\insertpaperteaser
\input{\PaperRoot src/related}
\input{\PaperRoot src/prelim}

\input{\PaperRoot src/method}
\input{\PaperRoot src/exp}

\FloatBarrier
\input{\PaperRoot src/conclusion}
\clearpage
\bibliographystyle{\PaperRoot ACM-Reference-Format}
\bibliography{\PaperRoot iclr2027_conference}
\clearpage
\appendix
\input{\PaperRoot src/appendix}

\end{document}

%% file: preamble.tex
% Scientific packages and annotations retained from the original manuscript.
\usepackage{nicefrac,multirow,tabularx,hhline,xspace,enumitem,pifont,float,longtable,placeins}
\input{\PaperRoot math_commands.tex}

\definecolor{darkgreen}{rgb}{0.0,0.5,0.0}

\definecolor{bestgreen}{HTML}{9FD3AA}
\definecolor{secondgreen}{HTML}{C6E5CC}
\definecolor{thirdgreen}{HTML}{E8F4EB}

%% file: figures/fig_teaser.tex
\centering
    \includegraphics[width=\linewidth]{\PaperRoot 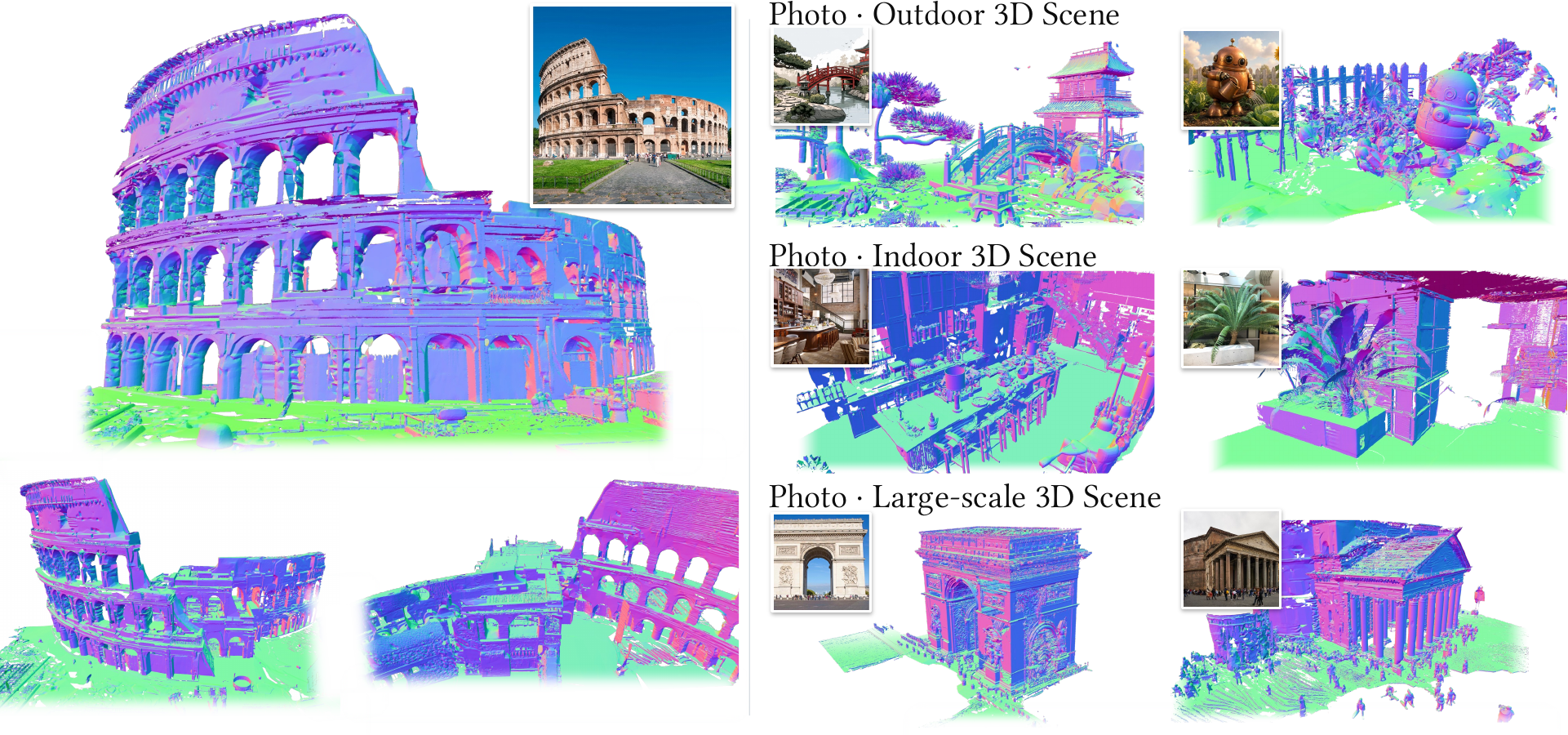}
    \caption{\textbf{Generating a complete 3D mesh from a single image.}
Our method reconstructs detailed scene meshes from one photograph. The method completes geometry beyond the observed view and supports indoor, outdoor, and large-scale scenes. We show the Colosseum reconstruction from multiple viewpoints (left) and diverse reconstruction results (right).}
\label{fig:teaser}

%% file: src/abstract.tex
Single-image scene generation aims to produce a complete 3D scene mesh from a single image, including surfaces the camera did not observe. While pretrained 3D object generators encode a strong shape prior, they are mainly designed for isolated objects in a fixed canonical volume and focus mostly on indoor scenes, since diverse 3D data for outdoor scenes are quite limited. In this work, we present a method that redesigns such an object-centric generator, \textit{e.g.}, Trellis~2, to work on both indoor and outdoor scenes while retaining its prior. We accomplish this by (a) partitioning the scene into adaptive chunks that scale relative to the distance to the camera; nearby chunks have a smaller size to keep the finer detail, while distant structures, \textit{e.g.}, buildings, are covered by large chunks; (b) making the generator capture explicit 2D--3D correspondence by lifting image features and making the model aware of the free space, observed surface, and unobserved region; (c) synthesizing around 4,000 outdoor scenes to broaden the training data, as existing scene datasets are largely indoor. Experiments on Tanks and Temples, ScanNet++, and in-the-wild images show that our method outperforms all baselines in geometric accuracy and perceptual quality across both indoor and outdoor scenes. 

%% file: src/intro_new.tex
\section{Introduction}

We propose a method for generating a complete 3D scene mesh from a single image. This is a long-standing problem in computer vision and graphics with applications in content creation, virtual reality, and robotics. This task has gained new relevance with world models that turn a single image into an explorable 3D world~\citep{hy2026hy, shen2026lyra}, reducing the need for extensive capture. Such worlds can be used in a physics simulator for robot navigation and interaction~\citep{shen2026lyra} to scale up robot learning~\citep{robocasa2024,robocasa365}. Different from multi-view reconstruction, generating a scene from a single image requires not only estimating the depth and shape of the visible surfaces, but also completing the geometry unobserved behind them. 

Recent approaches to single-image scene generation leverage 2D generative priors by distilling into a 3D representation~\citep{sargent2024zeronvs}, iteratively inpainting and lifting novel views~\citep{hollein2023text2room}, or generating views or videos that are then reconstructed into 3D~\citep{yu2024viewcrafter, gao2024cat3d, shen2026lyra}. Feedforward 3D generators~\citep{xiang2025structured, zhao2025hunyuan3d} trained on large object collections have been extended to indoor scenes by composing per-object generations~\citep{huang2025midi, meng2026scenegen} or by predicting scene-level geometry directly~\citep{schmid2026genrecon, zhang2026world}. However, learning a scene-level generative prior remains difficult as diverse 3D scene data are limited. Unlike the large asset collections behind object generators~\citep{deitke2023objaverse, zhang2025texverse}, scene datasets~\citep{dai2017scannet, fu20213d, xia2026sage} focus on indoor scenes, and procedural generators~\citep{raistrick2023infinite} where the outdoor scenes' diversity is limited by hand-crafted rules and asset generators. Training a model that handles the layouts, structures, and viewpoints of general scenes therefore remains difficult.

We present a novel method building upon Trellis~2~\citep{xiang2026native} for scene reconstruction while inheriting its learned prior on objects. To better handle this task, our method redesigns Trellis~2 by explicitly imposing 2D--3D correspondences using a monocular point map as geometric conditioning. This raises three technical challenges: (a) representing scenes whose physical extent far exceeds an object's canonical volume, (b) keeping the generated geometry faithful to the observed image, and (c) training with limited outdoor scene data. We address these as follows.

First, we represent a scene with \emph{adaptive chunks} whose physical scale varies across the scene. Fitting an entire scene into the generator's fixed-resolution volume sacrifices detail, while fixed-size chunks~\citep{schmid2026genrecon} require many generations to cover a large scene. Under perspective projection, distant regions occupy fewer pixels, so we use smaller chunks nearby to preserve detail and larger chunks for distant regions to extend coverage. Second, we condition each 3D voxel on the image through explicit 2D--3D correspondence. Using an estimated point map, we lift image features~\citep{simeoni2025dinov3} only onto the observed surface and encode whether each token lies in free space in front of the surface, on it, or in the unknown region behind it. This indicates to the model where to reconstruct and where to generate. 
Third, we construct around 4,000 diverse outdoor scenes with a synthetic pipeline and train on them together with existing indoor datasets. While the outdoor scenes are not perfect reconstructions, we found, somewhat surprisingly, that training on them improves the model's performance.

Empirically, we achieve strong performance in single-image reconstruction over baselines. \figref{fig:teaser} shows reconstructions of indoor, outdoor, and large-scale scenes, including the Colosseum viewed from multiple viewpoints. We evaluate against baselines on Tanks and Temples~\citep{tanks}, ScanNet++~\citep{yeshwanthliu2023scannetpp}, and 121 in-the-wild images, and our method outperforms all baselines on every metric. On Tanks and Temples, it reduces Chamfer distance by 24\% and improves F1 by 30\% over the strongest baseline on each metric, and on in-the-wild images it is preferred over every baseline in at least 73\% of pairwise user-study comparisons. Our adaptive chunking also runs three times faster than fixed-size chunking while improving reconstruction quality.

{\bf Our main contributions are:}
\begin{itemize}[topsep=0pt, leftmargin=12pt]
    \setlength{\itemsep}{2pt}
    \setlength{\parskip}{2.5pt}
    \item {\bf Adaptive scene chunks.} We introduce a chunk-based scene representation whose physical scale changes across the scene. Nearby regions use smaller chunks to preserve detail, while farther or taller regions use larger chunks to extend scene coverage, so that one generator trained on a canonical volume can cover scenes of varying extent.
    \item {\bf Explicit 2D--3D conditioning.} We propose a scene mesh generation model that explicitly distinguishes 3D locations supported by the observation from unobserved regions, allowing reconstruction and generative completion to be handled within a unified model.
    \item {\bf Outdoor scene supervision.} We construct around 4,000 diverse outdoor scenes to create additional training supervision. The model trained on them produces better reconstructions at inference time than the scene-generation framework used to create the dataset.
\end{itemize}

%% file: src/related.tex
\section{Related Work}
\input{\PaperRoot figures/fig_pipeline}

{\bf Scene generation with image and video priors.} A line of work reconstructs 3D scenes by leveraging pretrained 2D generative models. \emph{Diffusion distillation} optimizes a 3D representation under image-generation priors~\citep{sargent2024zeronvs, poole2022dreamfusion}, enabling novel-view synthesis from limited observations, but it requires costly per-scene optimization. \emph{Iterative lifting and inpainting} instead alternates between rendering the current reconstruction, synthesizing unobserved regions, and lifting them back into 3D~\citep{hollein2023text2room}. Such pipelines can progressively expand a scene, but errors in inpainting and geometric registration may accumulate across iterations ~\citep{li2026rgb}.
More recent \emph{generate-then-reconstruct} approaches generate multi-view images and reconstruct them into a 3D scene~\citep{yu2024viewcrafter, gao2024cat3d, shen2026lyra, hy2026hy}, which could improve consistency, but reconstruction quality depends on both cross-view consistency and whether the chosen viewpoints sufficiently reveal occluded geometry.
In contrast, our method predicts scene geometry directly from the input image without first synthesizing a set of novel views.

{\bf Compositional reconstruction.}
Large-scale object datasets have enabled strong feed-forward 3D generators~\citep{xiang2025structured,zhao2025hunyuan3d,li2025step1x}. These models learn
useful complete-shape priors, but are designed for isolated assets in a canonical
volume. Compositional methods extend them to scenes by decomposing an image into
objects, generating or retrieving each object independently, and estimating their
layout~\citep{zadaianchuk2026reconstruction,huang2025midi,meng2026scenegen}.
This formulation works naturally for separable instances, but is less suited to
continuous scene structures such as floors and walls. Training-free methods such
as EvoScene and Extend3D instead reuse object generators through local generation
and inference-time refinement~\citep{zheng2026self,yoon2026extend3d}, requiring
new geometry to be repeatedly reconciled with the existing scene. We instead
learn scene-level consistency from training, enabling more robust generation.

{\bf Scene-level generation.} Recent methods directly model scenes rather than composing independent objects. GenRecon~\citep{schmid2026genrecon} extends an object prior with overlapping scene chunks, but uses a shared physical scale determined by scene height, creating a tradeoff between scene coverage and local resolution. We instead vary chunk scale across the scene to preserve nearby detail and cover larger structures. Other methods focus on how scene geometry is represented or conditioned. RecGen~\citep{zadaianchuk2026reconstruction} integrates point maps into image features, SAM~3D~\citep{chen2026sam} uses geometric tokens, while layered, point-based, and volumetric methods directly predict visible and hidden scene geometry~\citep{li2025lari,zhang2026world,chen2026nova3r,ngo2026volfill}. Differently, we exploit the spatial locations of the generator's 3D tokens to directly condition on the image features, % with the geometry it constrains, 
while completing occluded parts in the mesh latent space.

{\bf 3D scene datasets.} 3D scene datasets are much smaller and less diverse than the assets used to train object generators. Existing sources include indoor scans and synthetic room datasets such as ScanNet, 3D-FRONT, and SAGE-10k~\citep{dai2017scannet,yeshwanthliu2023scannetpp,fu20213d,xia2026sage}, while Infinigen provides procedurally generated natural environments~\citep{raistrick2024infinigen}. However, outdoor scenes have much greater variation in geometry, scale, and layout. We complement these datasets with outdoor scenes constructed using an agentic framework.

%% file: figures/fig_pipeline.tex
\begin{figure}[t]
\small
    \centering
\includegraphics[width=0.99\linewidth]{\PaperRoot 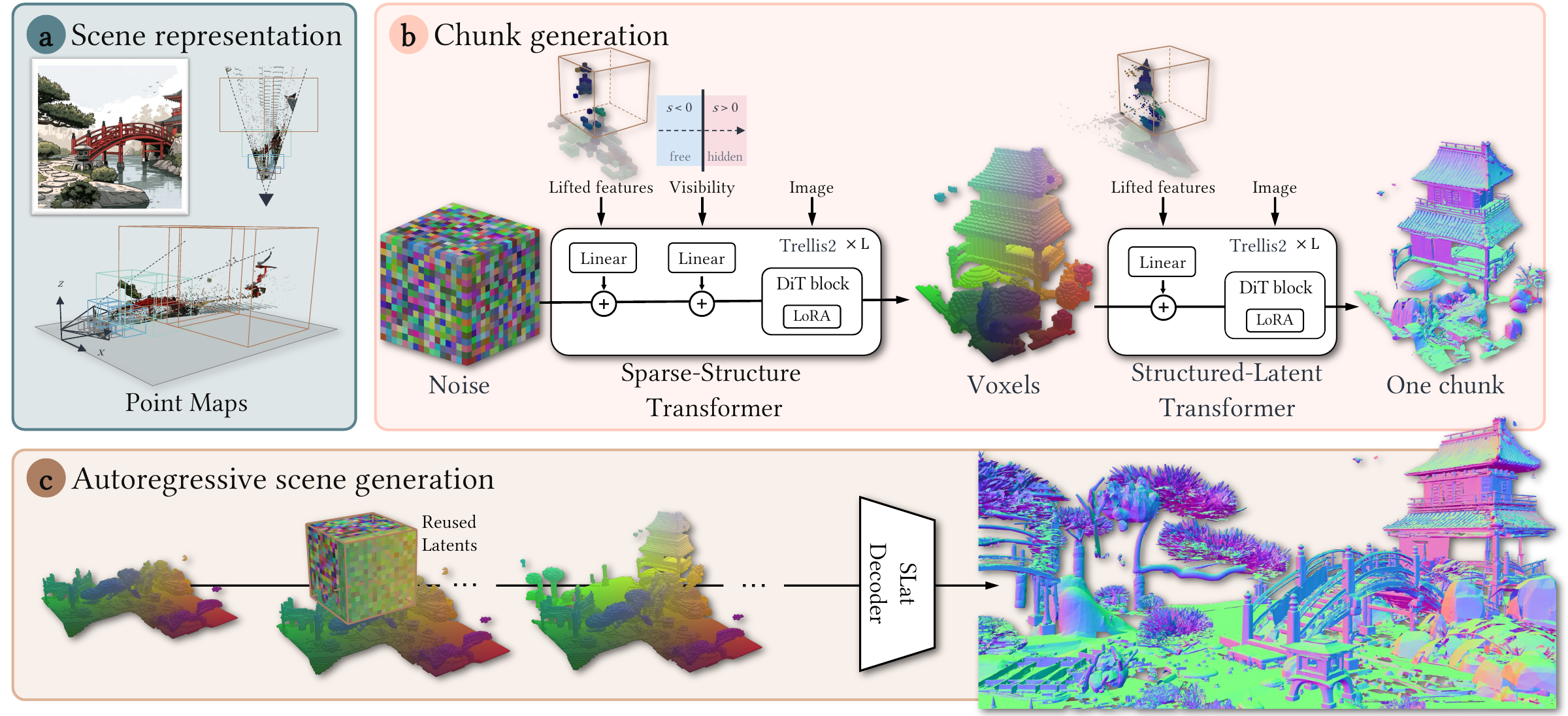}
\caption{\textbf{Overview of our method.} \emph{(a) Scene representation.} We lift the input image using an estimated point map and partition the scene into adaptive 3D chunks. \emph{(b) Chunk generation.} For each chunk, %our two-stage Trellis~2-based 
the generator first predicts the sparse structure and then its structured geometry latents. Both stages are conditioned on surface-aligned lifted image features, while the first stage further receives a depth-based visibility indicator to distinguish free space vs. occluded regions.
%
%conditioned on the input image,  DINO features lifted near the observed surface, and a visibility indicating free space or in the unobserved region. 
%
\emph{(c) Autoregressive scene generation.} We generate neighboring chunks sequentially while reusing latents in overlapping regions, allowing each new chunk to complete unseen geometry conditioned on the previously generated scene chunk. The assembled latents are finally decoded into the full scene mesh.}
\label{fig:pipeline}
\end{figure}

%% file: src/prelim.tex
\section{Background}
Trellis~2~\citep{xiang2026native} is an image-conditioned 3D asset generator trained on \textbf{object-centric data}. Its geometry representation is based on O-Voxels, which store surface information at active voxel locations, while empty voxels are discarded. This representation does not require watertight geometry and supports open surfaces and interior surfaces, making it suitable for scene structures. A VAE encodes O-Voxel features into structured
latents associated with explicit 3D voxel locations. Geometry generation proceeds through two flow-matching models.
The sparse-structure stage generates a coarse latent grid that is
decoded into active voxel locations. The structured-latent stage
then generates geometry latents at these locations, which the
pretrained decoder converts into a mesh. Both stages operate within
the normalized volume $[-0.5, 0.5]^3$. However, Trellis~2 is trained to generate objects centered and normalized within a canonical volume, rather than recover their scale and layout in a scene. Extending it to scenes therefore requires preserving layout. Fitting the entire scene into one fixed-resolution volume would lose detail. This motivates adapting both its spatial representation and conditioning for scene reconstruction.

%% file: src/method.tex
\section{Scene Mesh Generation}

{\bf Problem formulation and challenges.} Given a single image, we aim to generate a scene mesh that reconstructs visible surfaces \textit{and} completes plausible geometry in occluded regions. Scene-level generation requires allocating resolution across large spatial extents, aligning generated geometry with the input observation, and learning from limited scene supervision. These requirements motivate the following three design choices:

\begin{enumerate}[topsep=0pt, leftmargin=16pt,label=(\alph*)]
    \setlength{\itemsep}{2pt}
    \setlength{\parskip}{2.5pt}
    \item Scenes span a larger range of physical sizes than individual objects. Representing an entire scene in a single fixed-resolution volume sacrifices local detail as the scene’s physical extent increases.
    One could also divide up the scene and scale each chunk to the canonical volume, but this would be inefficient as one would need a sufficiently fine-grained resolution. Hence, we propose \emph{adaptive chunks} that preserve finer resolution near the camera while increasing coverage where needed (\secref{sec:scene_rep}).
    \item Scene generation must remain faithful to the observed image content. As Trellis~2 is trained on centered objects, using it for scene generation leads to the entire scene being generated at the canonical volume's origin with incorrect layout. For scene generation, the model must place visible surfaces at their observed position and depth, and complete the unobserved geometry. To make this learning efficient, we provide explicit 2D--3D correspondence to the model by lifting image features onto the observed surfaces in a free-space-aware manner (\secref{sec:chunk_conditioning}).
    \item This work aims to generate outdoor scenes, but existing datasets offer limited coverage and diversity of such scenes. To broaden the training data, we synthesize outdoor 3D scenes and use them with existing indoor scene datasets to train our model (\secref{sec:training}).
\end{enumerate}

\subsection{Representing a scene with adaptive chunks}
\label{sec:scene_rep}
\input{\PaperRoot figures/fig_adapt}

A photo of a scene may contain objects that are close to the camera and ones further away, \eg, a nearby chair and a building far behind it. Representing both the chair and the building in a fixed volume makes the chair lose most of its details as the chair is scaled relative to the building (\figref{fig:adapt}a). One could also represent the scene using a fixed-resolution grid by dividing the scene into overlapping chunks, similar to GenRecon~\citep{schmid2026genrecon}. The overall scene volume is obtained by fitting a rectangular 3D bounding volume to an estimated point map, then splitting it into chunks. However, using the same physical chunk size throughout the scene still leaves a choice between many small predictions or fewer, coarser predictions (\figref{fig:adapt}b). 

{\bf Adaptive chunking.} We instead let the physical chunk size adapt based on the scene's geometry. Under perspective projection, a region of the same physical size occupies fewer pixels in the image as its distance from the camera increases. Nearby regions use smaller chunks to preserve local details, while distant regions use larger chunks to cover more space per generation, as shown in~\figref{fig:adapt}c. At inference time, we use the estimated point map to cover the scene from the camera. 

We begin with the smallest chunk size wherever the observed height permits. As coverage extends farther away, we double the physical chunk size and increase the scale when needed to accommodate tall structures. Within each region, adjacent chunks overlap by $0.5\,\mathrm{m}$ in canonical space, corresponding to a physical overlap of $0.5a_i\,\mathrm{m}$. Overlapping chunks provide shared geometry for autoregressive generation. This allocation preserves fine resolution near the camera where possible while allowing distant or tall regions to occupy larger chunks. Appendix~\ref{app:chunk} provides more detail and comparisons of chunking strategies. Our chunk allocation therefore depends on both depth and local height, allowing the scene's coverage to grow without forcing all regions to use the same resolution.

{\bf Supporting different physical scales in one model.} 
To use a single model across adaptive chunks, we need a consistent representation. This is done by expressing all geometry in a ground-aligned coordinate frame whose origin is the point on the ground plane directly below the camera center, with $+z$ upward, $+y$ following the camera-forward direction, and $+x$ the lateral axis. 

We divide the scene along $+y$ into overlapping \emph{depth regions},
each covering a range of forward distances from the camera. All chunks within depth region $i$ share a scale factor $a_i \geq 1$. The model always operates on a canonical chunk of side length $L=3\,\mathrm{m}$, while the physical extent is $L_i=a_iL$ and the voxel size in region $i$ is $\Delta_i = \frac{a_iL}{N}$, where $N$ is the number of voxels along each axis. Larger chunks cover more space with the same grid dimensions, while smaller chunks allocate more voxels to local geometric detail.

For a chunk in region $i$, we divide its 3D coordinates, depth values, and camera translation by $a_i$, while keeping the camera intrinsics unchanged. A physical cube of side length $a_iL$ is thus mapped to the same canonical cube of side length $L$. As the geometry and camera translation are scaled together, image projections remain unchanged, allowing the same model and 2D--3D conditioning mechanism to operate across chunks with different physical resolutions.

\subsection{Chunk Conditioning for Explicit 2D--3D Correspondence}
\label{sec:chunk_conditioning}

Adaptive chunks just determine \emph{where} the model generates and the physical resolution used in each region. To generate a chunk, the model would require knowing which parts of its 3D volume are visible in the input image. For a visible surface, the generated geometry should follow the observation. Behind it, the geometry is unobserved and thus should follow the generative prior. We obtain the visible geometric information by using the monocular point map estimated from the input image. 

To provide this geometric information to the model, one could consider image conditioning using cross-attention layers. For example, Trellis~2 lets every 3D token attend to the same image features, leaving the model to learn which image regions correspond to each 3D location. Geometry-aware approaches such as RecGen~\citep{zadaianchuk2026reconstruction} add point maps into image features, while SAM~3D~\citep{chen2026sam} introduces additional geometry tokens through cross-attention. 

However, these conditioning methods do not exploit that the position of every 3D token in adaptive chunking is already known. We therefore condition each token, via 2D–3D projection, on the image feature at its projected location and on its position along the camera ray relative to the observed surface.

For a voxel centered at $v$ in the normalized scene frame, let $\pi(v)$ be its image projection, and $z_{\mathrm{cam}}$ be its camera-space depth. We define the depth residual:
\begin{equation}
    r(v) = z_{\mathrm{cam}}(v) - D\bigl(\pi(v)\bigr),
    \label{eq:depth_residual}
\end{equation}
where $D$ denotes the observed camera-depth map after scale normalization. Sky pixels and invalid depths are excluded. Note, all quantities here are expressed in the canonical frame of the current chunk. We omit the region index $i$ for readability.

{\bf Lifting image features onto observed surfaces.}
From the input image, we first extract a feature map $F$ using DINOv3~\citep{simeoni2025dinov3}. Next, we assign this DINO feature $F$ to a voxel/token only when its center is close to
the observed surface along the viewing ray:
\begin{equation}
    f_{\mathrm{lift}}(v) =
    \begin{cases}
        F\bigl(\pi(v)\bigr),
            & |r(v)|<\tau, \\
        0, & \text{otherwise}.
    \end{cases}
    \label{eq:feature_lifting}
\end{equation}
The tolerance $\tau$ accounts for voxel centers that do not fall exactly on a
surface. We set $\tau$ to be $1.5$ voxel widths, giving $\tau = 0.28\,\mathrm{m}$ for the sparse-structure stage and $\tau=0.07\,\mathrm{m}$ for the structured-latent stage (in canonical units). In chunk depth region $i$, the physical tolerance is $a_i\tau$. 

{\bf Distinguishing free space from occlusion.} 
%A token
A voxel $v$ without a lifted feature can lie either in front of a visible surface or behind it. These locations require different behavior. The
former should remain empty, while the latter may contain hidden geometry. We provide this information to the sparse-structure model using the following clipped signed-depth residual
\begin{equation}
    s(v) = \operatorname{clamp}\!\left(r(v),-\delta,\delta\right),
    \qquad \delta=1\,\mathrm{m},
    \label{eq:depth_clearance}
\end{equation}
 measured in canonical units. Negative values indicate space between the camera and the visible surface, values near zero indicate the surface, and positive values indicate space behind it. We Fourier-encode~\citep{nichol2021glide} this scalar $s(v)$, followed by an MLP, \ie,
\begin{align}
    g(v)&=\operatorname{MLP}\!\left(\gamma(s(v))\right) \quad \text{and} \quad \gamma(s)
        =\bigl[\cos(\symbfit{\omega}s),
                 \sin(\symbfit{\omega}s)\bigr],
        \label{eq:geometry_encoding}
\end{align}
where $\symbfit{\omega}\in\mathbb{R}^{64}$ contains the fixed frequencies, giving a 128-dimensional embedding. Before each transformer block $l$, we project and add local conditions to the tokens. In other words, the sparse-structure model's hidden states $h$ are replaced as follows:
\begin{equation}
    h_\ell(v) \leftarrow h_\ell(v)
        +W^{\mathrm{img}}_\ell f_{\mathrm{lift}}(v)
        +W^{\mathrm{geom}}_\ell g(v),
    \label{eq:condition_injection}
\end{equation}
where $W^{\mathrm{img}}_\ell$ and $W^{\mathrm{geom}}_\ell$ denote learned
linear projections. The structured-latent model uses only the lifted-image term, since visibility and occupancy are resolved during sparse-structure generation.

\subsection{Training \& Generation}
\label{sec:training}
Training scene generation requires mesh ground-truth that includes the full geometry, even those not visible in the input image view. For this, we use SAGE-10k~\citep{xia2026sage}, which contains $10{,}000$ indoor scenes generated with each of Infinigen 1.0 and 2.0~\citep{raistrick2024infinigen}. For each scene, we render $16$ views at $1024\times1024$ resolution using Blender.
We keep target geometry within each camera frustum and discard objects occupying fewer than $80$ pixels.

{\bf Outdoor supervision geometry.} To get diverse outdoor supervision, we generate $4{,}000$ reference images from 20 outdoor categories using a proposed agentic framework. First, we use GPT-5.6~\citep{openai2026gpt56} to generate text prompts describing a scene, including the 
location of the nearby objects, terrain, and vegetation.
We then use Ideogram 4~\citep{ideogram-4-2026} to generate reference images from these prompts. For each image, a VLM identifies object types, estimates their dimensions, and specifies spatial relations between objects. We crop the objects, complete canonical views with
Nano Banana~\citep{google2025nanobananapro}, and convert them to
meshes with Trellis~2. 
Following SAGE-10k~\citep{xia2026sage}, a layout solver
converts the relations into object poses while rejecting collisions. Ground geometry combines guidance from a top view of the estimated
monocular geometry with procedural height fields~\citep{guo2026worldclaw}. We then render training images from the assembled meshes. This makes the input image, depth, camera, and target geometry consistent even when the assembled scene differs from its reference image. Although the assembled scenes may contain inaccurate shapes and layouts, they still provide useful supervision. As shown in~\figref{fig:imperfect},
our trained model produces more detailed geometry and better preserves
the input layout than the synthetic pipeline used to construct its training data. The full recipe is detailed in Appendix~\ref{app:agentic_framework}.

\input{\PaperRoot figures/fig_imperfect}{\bf Training on canonical chunks.} We train on individual canonical chunks so that local scene completion does not require processing an entire scene at once, which saves memory. During training, we normalize each scene to a common spatial range and extract overlapping chunks of side length $L$. Geometry, depth, and camera translation are scaled consistently, preserving their image projections. Ground-truth depth and camera parameters provide conditioning during training, while monocular estimates replace them at inference. We optimize rank-32 LoRA adapters~\citep{hu2021lora} together with the new conditioning modules. At inference, each adaptive region is rescaled by $1/a_i$ into the same canonical domain seen during training.

{\bf Autoregressive generation of a scene.}
Given an image, adaptive chunking determines where to generate. 
We then generate the chunks sequentially, starting with chunks closer to the camera.
For each chunk, we map its latent cells into the shared scene frame and identify those already covered by the interior of previously generated chunks. Their latent values are then copied into the new chunk and re-imposed at every denoising step, so overlapping chunks agree by construction. Across depth regions, whose latent cells are of different sizes, we trilinearly resample the previous region's latents at the new cell locations within the overlap and constrain them in the same way. We repeat this process over the allocated chunks and decode the assembled latents into the scene mesh.

%% file: figures/fig_adapt.tex
\begin{figure}[t]
  \centering
  \includegraphics[width=\linewidth]{\PaperRoot 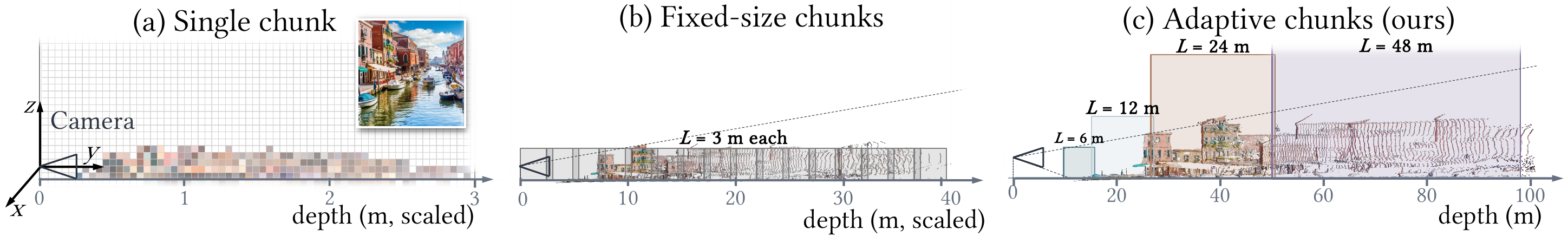}
  \caption{\textbf{Scene chunking strategies.}
\emph{A single chunk (a)} compresses the entire scene into one fixed volume, sacrificing geometric detail, while \emph{fixed-size chunks (b)} require many chunks to cover large scenes. \emph{Our adaptive strategy (c)} increases the physical chunk size with depth, preserving finer resolution nearby while efficiently covering distant regions. {\bf Best viewed zoomed in.}} 
  \label{fig:adapt}
\end{figure}

%% file: figures/fig_imperfect.tex
\begin{figure}[htbp]
\centering
\begin{minipage}{0.55\linewidth}
  \centering
  \hfill
  \includegraphics[width=.95\linewidth]{\PaperRoot 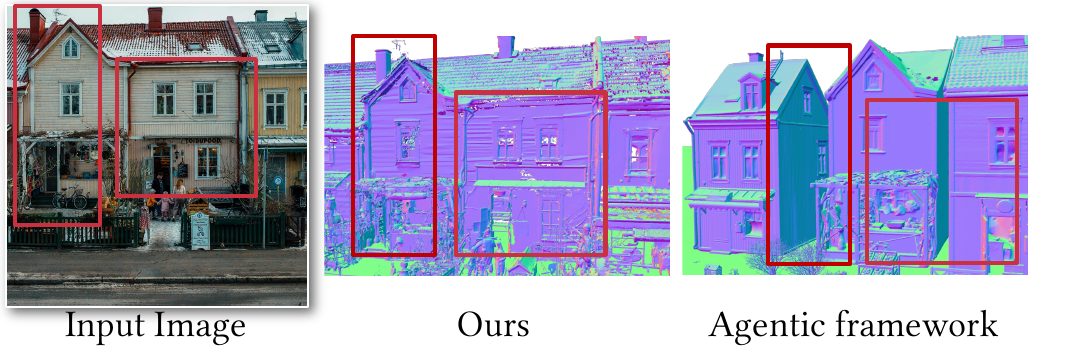}
  \caption{\textbf{Learning from imperfect synthetic data.} Our model reconstructs more faithful geometry than the imperfect training data from the agentic framework.}
  \label{fig:imperfect}
\end{minipage}
\end{figure}

%% file: src/exp.tex
\section{Experiments}
We evaluate our method against seven single-image 3D reconstruction baselines on real-world scans and in-the-wild images. Our experiments examine geometric accuracy and consistency with the input image. We then ablate each of our core design choices.

{\bf Datasets.} We train on the synthetic scene corpus described in \secref{sec:training}. For quantitative geometry evaluation, we use 7 scenes from Tanks and Temples~\citep{tanks} and 32 randomly selected scenes from ScanNet++~\citep{yeshwanthliu2023scannetpp}. We additionally collect 121 in-the-wild images from Atlas~\citep{worldlabs2026atlas}, Google Images of tourist attractions, NeRF~\citep{mildenhall2021nerf} and mip-NeRF~\citep{barron2022mip} datasets, and MoGe examples~\citep{wang2026moge, kong2026moge}, comprising 53 indoor and 68 outdoor scenes. %Some of our images are provided in Appendix~\ref{sec:more_results}. 
We use a separate validation set for the ablations.

\input{\PaperRoot figures/fig_visual_comp}
{\bf Implementation details.} We initialize our model from Trellis~2 and train it for 100K steps using AdamW~\citep{loshchilov2017decoupled} on 16 NVIDIA A100 GPUs with a batch size of 64. We use learning rates of $1 \times10^{-4}$ for the newly introduced conditioning modules and $3\times10^{-5}$ for LoRA~\citep{hu2021lora}. During inference, we use MoGe-3~\citep{kong2026moge} to predict camera intrinsics and point maps. Technical architectural and training details are provided in~\secref{sec:chunk_conditioning} and~\secref{sec:training}.

{\bf Baselines.} We compare our method against four categories of single-image 3D scene reconstruction approaches: \emph{iterative 3D completion} (EvoScene~\citep{zheng2026self} and Extend3D~\citep{yoon2026extend3d}), \emph{object composition} (3D-RE-GEN~\citep{sautter20263d}), \emph{generated video followed by reconstruction} (Lyra~2.0~\citep{shen2026lyra}), and \emph{direct
scene reconstruction} (VolFill~\citep{ngo2026volfill}, GenRecon~\citep{schmid2026genrecon}, and World Tracing~\citep{zhang2026world}). All methods are evaluated with a single conditioning RGB image. For Lyra~2.0, which requires a camera trajectory, we follow the CAT3D camera
heuristics~\citep{gao2024cat3d}: a forward-moving spiral for scene-centric images and a circular trajectory for object-centric images. 

{\bf Metrics.}
For datasets with ground-truth 3D geometry, including Tanks and Temples~\citep{tanks} and ScanNet++~\citep{yeshwanthliu2023scannetpp}, we evaluate geometric reconstruction using Chamfer Distance (CD) and F1 score. We represent both predictions and ground truth as point clouds. For methods that predict point clouds, we use the predicted points directly. For methods that predict meshes, we sample points from the surface using area-weighted sampling. For Lyra~2.0, we use the Gaussian centers as the predicted points. We similarly convert the ground-truth geometry to point clouds. Before computing the metrics, we use ICP to align each prediction with the ground truth. CD measures the geometric discrepancy between the prediction and ground truth. To make CD comparable across scenes with different scales, we normalize it by the median ground-truth depth. For F1, we use a distance threshold of $10\,\mathrm{cm}$, where a predicted point is considered correct if it lies within $10\,\mathrm{cm}$ of the ground-truth surface.

For in-the-wild images without ground-truth 3D geometry, we instead evaluate consistency with the input image using DreamSim~\citep{fu2023dreamsim} and CLIP-N~\citep{radford2021learning}. DreamSim measures perceptual similarity and correlates with human judgments. CLIP-N measures geometric consistency by comparing rendered normal maps with the normal map estimated from the input image using Lotus2~\citep{he2025lotus}. We render the reconstructed scene from four viewpoints using surface normals and compute the similarity between the rendered and estimated normal maps. We conduct a user study using blind pairwise comparisons. Participants are shown an input image together with two reconstructed 3D scenes and asked to select the better reconstruction. 
%which reconstruction better matches the input image.

\subsection{Comparisons}
\tabref{tab:quantitative_comparison} compares ours against baselines on Tanks and Temples~\citep{tanks}, ScanNet++~\citep{yeshwanthliu2023scannetpp}, and in-the-wild images. Our approach outperforms all baselines across every metric and user study. In~\figref{fig:visual_comp}, we show some qualitative results. For additional results, please see \figref{fig:more_results} and \figref{fig:more_results2} in the appendix. GenRecon~\citep{schmid2026genrecon} often produces room-like structures in outdoor scenes, suggesting that its learned scene prior can dominate when the input differs from its training data. World Tracing~\citep{zhang2026world} often leaves holes under heavy occlusion, indicating difficulty completing hidden regions while maintaining the scene structure. We also find that generating realistic views does not necessarily yield a coherent scene. Lyra~2.0~\citep{shen2026lyra} struggles with inconsistencies that lead to floating geometry after reconstruction. 

\input{\PaperRoot src/table/table1}
\input{\PaperRoot src/table/table2}
\subsection{Ablation Studies}
{\bf Adaptive chunks.}
We compare three chunking strategies using the same trained model:
normalizing the entire scene to fit within a single chunk;
normalizing the scene height to fit within 3\,m and using fixed 3\,m chunks~\citep{schmid2026genrecon};
and our adaptive strategy, which uses smaller chunks nearby and larger
chunks farther away.
As shown in~\tabref{tab:layout_ablation}, adaptive chunking achieves the
best reconstruction scores on both indoor and outdoor scenes while
using fewer chunks and running three times faster than fixed-size chunking.
The single-chunk strategy is competitive indoors but performs
substantially worse outdoors, consistent with the loss of local detail
when a large scene is compressed into a fixed-resolution volume. Fixed-size chunks sometimes produce duplicated objects and incorrect layouts.
We suspect that small chunks covering only fragments of the scene provide insufficient spatial context for scene
completion. Larger distant chunks can capture more of these structures
within one generation, while smaller nearby chunks preserve local detail.

{\bf Conditioning mechanisms.}  We ablate different conditioning mechanisms in \tabref{tab:cond_mech}. \emph{SAM~3D-style conditioning}~\citep{chen2026sam} tokenizes the point maps and supplies them as additional tokens alongside the image tokens. \emph{Projection-only conditioning} directly unprojects image features into 3D, similar to a plane-sweep volume. Our conditioning outperforms these alternatives on both the indoor and outdoor validation sets. From inspecting the reconstructed meshes, we observe that projection-only and SAM~3D-style conditioning more often produce duplicated surfaces and miss geometry in occluded regions. We hypothesize that explicitly associating each 3D token with both its corresponding image feature and its position relative to the observed surface makes the model easier to learn. The model no longer needs to implicitly infer the 2D--3D relationships through cross-attention or unprojected image features alone.

{\bf Contribution of outdoor data.} We examine the contribution of our synthetic outdoor data by keeping our
conditioning mechanism fixed and comparing training on SAGE-10k
and Infinigen with and without our
outdoor data. As shown in~\tabref{tab:data_ablation}, adding outdoor
data improves reconstruction on both indoor and outdoor scenes,
with larger gains for outdoors. Meanwhile,
\tabref{tab:cond_mech} shows that our conditioning
outperforms other conditioning when
trained on the same data. These experiments indicate
that the improvements come from both our conditioning
and the additional data.

%% file: figures/fig_visual_comp.tex
\begin{figure}[t]
\small
    \centering
\includegraphics[width=\linewidth]{\PaperRoot 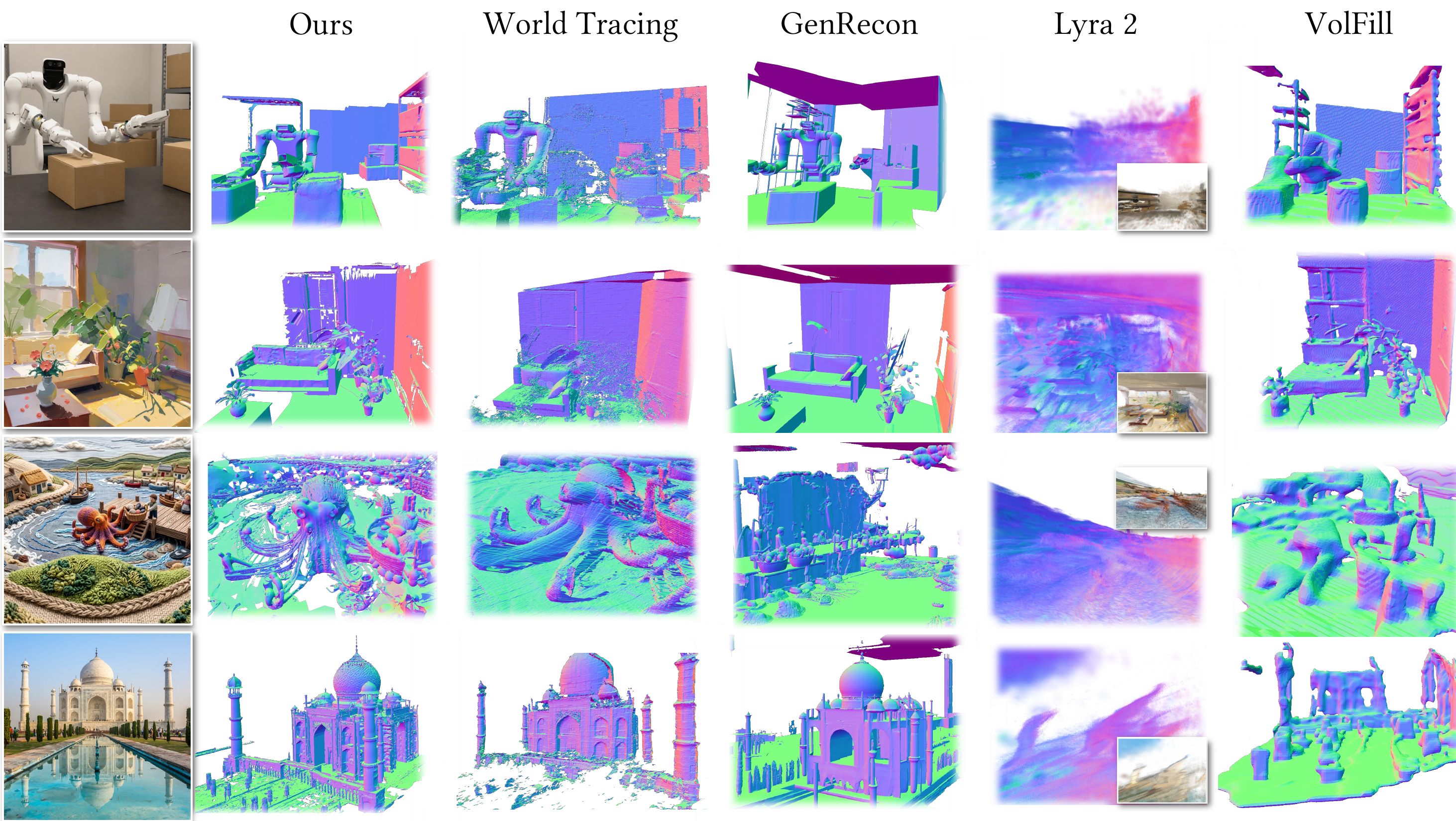}
\caption{\textbf{Visual comparisons.} Our method preserves global scene layout while recovering finer details across diverse inputs.
World Tracing~\citep{zhang2026world} produces noisy occluded regions and lacks detail.
GenRecon~\citep{schmid2026genrecon} often imposes a room-like prior for outdoor scenes. For Lyra 2~\citep{shen2026lyra}, we overlay RGB on the normal map. Although the rendering appears plausible, the underlying geometry is inaccurate.
VolFill~\citep{ngo2026volfill} has line-like surface artifacts and fails on outdoor scenes.}
\label{fig:visual_comp}
\end{figure}

%% file: src/table/table1.tex
\begin{table*}[t]
\centering
\small
\setlength{\tabcolsep}{3pt}
\caption{Quantitative comparison on Tanks and Temples, ScanNet++ and In-the-wild images. Our approach outperforms all baselines across different datasets and metrics.}
\label{tab:quantitative_comparison}
\label{tab:geometry_comparison}
\label{tab:outdoor_wild}
\resizebox{0.95\textwidth}{!}{
\begin{tabular}{lccccccc}
\toprule
& \multicolumn{2}{c}{\textbf{Tanks and Temples}}
& \multicolumn{2}{c}{\textbf{ScanNet++}}
& \multicolumn{3}{c}{\textbf{In-the-wild}} \\
\cmidrule(lr){2-3}
\cmidrule(lr){4-5}
\cmidrule(lr){6-8}

\textbf{Method}
& \textbf{CD} $\downarrow$ & \textbf{F1} $\uparrow$
& \textbf{CD} $\downarrow$ & \textbf{F1} $\uparrow$
& \textbf{DreamSim} $\downarrow$ & \textbf{CLIP-N} $\uparrow$
& \textbf{Win rate} (\%) $\uparrow$ \\
\midrule

\multicolumn{8}{l}{\textit{Iterative 3D completion}} \\
EvoScene~\citep{zheng2026self}
& 4.61
& 0.241
& 5.70
& \cellcolor{secondgreen} 0.690
& 0.379
& 0.797
& \textbf{90.7} \\

Extend3D~\citep{yoon2026extend3d}
& 8.37
& 0.165
& \cellcolor{thirdgreen} 5.54
& \cellcolor{thirdgreen} 0.674
& 0.375
& 0.805
& \textbf{94.4} \\

\midrule
\multicolumn{8}{l}{\textit{Object composition}} \\
3D-RE-GEN~\citep{sautter20263d}
& 7.17
& 0.146
& 15.84
& 0.291
& 0.429
& 0.810
& \textbf{88.9} \\

\midrule
\multicolumn{8}{l}{\textit{Generated views/video $\rightarrow$ 3D reconstruction}} \\
Lyra 2.0~\citep{shen2026lyra}
& \cellcolor{thirdgreen} 3.05
& \cellcolor{secondgreen} 0.388
& 8.17
& 0.481
& 0.347
& 0.810
& \textbf{94.3} \\

\midrule
\multicolumn{8}{l}{\textit{Direct scene reconstruction}} \\
VolFill~\citep{ngo2026volfill}
&\cellcolor{secondgreen} 2.63
& \cellcolor{thirdgreen} 0.376
& --
& --
& 0.339
& 0.789
& \textbf{89.7} \\

GenRecon~\citep{schmid2026genrecon}
& 5.57
& 0.215
& 10.06
& 0.411
& \cellcolor{thirdgreen} 0.294
& \cellcolor{thirdgreen} 0.847
& \textbf{73.0} \\

World Tracing~\citep{zhang2026world}
& 3.27
&  0.337
& \cellcolor{secondgreen} 5.21
& \cellcolor{secondgreen} 0.690
& \cellcolor{secondgreen} 0.237
& \cellcolor{secondgreen} 0.859
& \textbf{90.4} \\

\textbf{Ours}
& \cellcolor{bestgreen} \textbf{1.99}
& \cellcolor{bestgreen} \textbf{0.506}
& \cellcolor{bestgreen} \textbf{5.01}
& \cellcolor{bestgreen} \textbf{0.702}
& \cellcolor{bestgreen} \textbf{0.221}
& \cellcolor{bestgreen} \textbf{0.870}
& Ref. \\

\bottomrule
\end{tabular}}
\end{table*}

%% file: src/table/table2.tex
\begin{table}[htbp]
\centering
\small
\setlength{\tabcolsep}{5pt}
\caption{Comparison of chunking strategies.
Adaptive chunking improves reconstruction and is faster than fixed-size chunks.}
\label{tab:layout_ablation}
\begin{tabular}[t]{lccccc}
\toprule
\multirow{2}{*}{\textbf{Layout}}
& \multicolumn{2}{c}{\textbf{Indoor}}
& \multicolumn{2}{c}{\textbf{Outdoor}}
& \textbf{Time} \\
\cmidrule(lr){2-3}\cmidrule(lr){4-5}
& \textbf{CD}$\downarrow$ & \textbf{F1}$\uparrow$
& \textbf{CD}$\downarrow$ & \textbf{F1}$\uparrow$
& \textbf{(min)}$\downarrow$ \\
\midrule
Single chunk      & 2.87 & 0.759 & 49.9 & 0.138 & \textbf{1} \\
Fixed 3m          & 3.13 & 0.782 & 38.5 & 0.100 & 15 \\
\midrule
\textbf{Adaptive} & \textbf{2.86} & \textbf{0.792} & \textbf{37.4} & \textbf{0.240} & 5 \\
\bottomrule
\end{tabular}
\end{table}

\begin{table}[htbp]
\centering
\small
\setlength{\tabcolsep}{5pt}
\caption{Ablation of conditioning mechanisms.
Our conditioning outperforms implicit conditioning alternatives.}
\label{tab:cond_mech}
\begin{tabular}[t]{lcccc}
\toprule
\multirow{2}{*}{\textbf{Variant}}
& \multicolumn{2}{c}{\textbf{Indoor}}
& \multicolumn{2}{c}{\textbf{Outdoor}} \\
\cmidrule(lr){2-3}\cmidrule(lr){4-5}
& \textbf{CD}$\downarrow$ & \textbf{F1}$\uparrow$
& \textbf{CD}$\downarrow$ & \textbf{F1}$\uparrow$ \\
\midrule
SAM~3D style              & 4.38 & 0.707 & 8.30 & 0.257 \\
Projection only           & 4.74 & 0.739 & 8.26 & 0.330 \\
Ours w/o visibility       & 4.24 & 0.757 & \textbf{6.60} & 0.464 \\
\midrule
\textbf{Our conditioning} & \textbf{3.54} & \textbf{0.793} & 6.90 & \textbf{0.501} \\
\bottomrule
\end{tabular}
\end{table}

\begin{table}[htbp]
\centering
\small
\setlength{\tabcolsep}{5pt}
\caption{Ablation of dataset contribution.
Our outdoor data improves generalization without sacrificing indoor quality.}
\label{tab:data_ablation}
\begin{tabular}[t]{lcccc}
\toprule
\multirow{2}{*}{\textbf{Variant}}
& \multicolumn{2}{c}{\textbf{Indoor}}
& \multicolumn{2}{c}{\textbf{Outdoor}} \\
\cmidrule(lr){2-3}\cmidrule(lr){4-5}
& \textbf{CD}$\downarrow$ & \textbf{F1}$\uparrow$
& \textbf{CD}$\downarrow$ & \textbf{F1}$\uparrow$ \\
\midrule
Existing data                 & 3.99 & 0.784 & 10.75 & 0.317 \\
\midrule
\textbf{Existing data + Ours} & \textbf{3.54} & \textbf{0.793} & \textbf{6.90} & \textbf{0.501} \\
\bottomrule
\end{tabular}
\end{table}

%% file: src/conclusion.tex
\section{Conclusion}
We proposed a single-image scene mesh generation method to reconstruct indoor and outdoor scenes. Our method introduces adaptive scene chunks, explicit 2D–3D conditioning, and supervision from roughly 4,000 synthesized outdoor scenes to reconstruct observed surfaces and complete hidden geometry across indoor and outdoor environments. Experiments show that our method outperforms all baselines on indoor, outdoor, and in-the-wild scenes. A detailed ablation study validates the necessity of the proposed designs. 
For future work, we plan to explore textured outputs, \eg, 3D Gaussian splatting, finer geometric detail, and supporting more than one input image.

%% file: src/appendix.tex
\setcounter{table}{0}
\setcounter{figure}{0}
\renewcommand{\thetable}{A\arabic{table}}
\renewcommand\thefigure{A\arabic{figure}}
\renewcommand{\theHtable}{A.Tab.\arabic{table}}%<---!!!!---
\renewcommand{\theHfigure}{A.Abb.\arabic{figure}}%<---!!!!---
\renewcommand\theequation{A\arabic{equation}}
\renewcommand{\theHequation}{A.Abb.\arabic{equation}}%<---!!!!---

\section{Appendix}

Please also see videos in the supplemental materials for qualitative comparisons.

\subsection{Chunk Allocation Details}
\label{app:chunk}
\input{\PaperRoot figures/fig_app_chunk_allo}

Given an input image, we first use MoGE-3~\citep{kong2026moge} to estimate the camera intrinsics and a 3D point map. We use these estimates to determine the scene's physical size and placement of the chunks.
We define the scene coordinate frame by placing the origin at the ground-plane projection of the camera center, with $+y$ pointing in the camera-forward direction and $+z$ pointing upward. Following GenRecon~\citep{schmid2026genrecon}, we estimate the ground height using the $2$nd percentile of the point-cloud heights. We then construct a grid of overlapping chunks along both the camera-forward and lateral directions. For each depth row, we place enough chunks laterally to cover the corresponding range of the estimated point cloud, with neighboring chunks overlapping by $0.5\,\mathrm{m}$ within the camera frustum. Consecutive depth rows also overlap by $0.5\,\mathrm{m}$.

Rather than globally rescaling the entire scene, we determine the chunk scale separately for each depth range. For a given range, if its $99$th-percentile height exceeds $2.8\,\mathrm{m}$, we increase the physical chunk size so that the corresponding geometry fits within the canonical $3\,\mathrm{m}$ volume. Thus, only regions requiring additional spatial coverage are generated at a coarser physical resolution. The chunk size also increases progressively with depth: each new row is at least twice the size of the previous row, resulting in physical chunk sizes of $6$, $12$, $24\,\mathrm{m}$, and so on. The height criterion may further increase the scale when necessary, as shown in \figref{fig:app_chunk}. We continue adding rows until the grid reaches the $99$th-percentile depth of the estimated point cloud.

\input{\PaperRoot figures/fig_app_chunk_comparison}

{\bf Chunking strategies comparison.} Here, we compare the different chunking strategies in \figref{fig:app_chunk_output} on a large-scale scene extending to $280\,\mathrm{m}$ in depth and $10\,\mathrm{m}$ in height. Using fixed-size chunks at a uniform scale would require roughly 200 chunks, exceeding the memory capacity of an NVIDIA A100 (80GB). In contrast, our adaptive strategy covers the scene with only four depth regions and 15 chunks, requiring approximately 5 minutes of inference. Peak GPU memory usage is $26$\,GB, while generating a single chunk requires approximately $8$\,GB. Despite the reduced number of chunks, our method preserves fine geometry in nearby structures, such as the boats, while still covering large buildings in the distance. The single-chunk baseline runs in seconds, but compresses the entire scene into one canonical volume and consequently fails to recover meaningful scene geometry.

\subsection{More results}
\label{sec:more_results}
 We show surface-normal renderings on diverse indoor and outdoor scenes in \figref{fig:more_results} and \figref{fig:more_results2}. Across challenging inputs containing people, vegetation, vehicles, animals,
    and large structures, our method better preserves the overall scene layout
    while recovering finer geometric details and occlusion. World Tracing~\citep{worldlabs2026atlas} often produces
    noisy or incomplete geometry, particularly around occlusions and object
    boundaries. GenRecon~\citep{schmid2026genrecon} can impose simplified or room-like structures that are
    inconsistent with outdoor scenes. For Lyra~2~\citep{shen2026lyra}, the RGB output appears visually plausible, but the underlying geometry is often noisy and distorted, likely because of inconsistencies across the generated video. VolFill~\citep{ngo2026volfill} frequently fails on outdoor scenes and often produces only coarse scene geometry.

\input{\PaperRoot figures/more_results}
\input{\PaperRoot figures/more_results2}

\subsection{Limitations}
Our method generates scene geometry without textures. In addition, Trellis~2 does not guarantee watertight meshes, and our model inherits this limitation, occasionally producing holes that require additional mesh processing. Our method also relies on estimated depth and camera intrinsics, whose errors can affect chunk placement, scene scale, and feature alignment. Our autoregressive generation may propagate errors from earlier chunks to later ones through reused overlap latents.

\subsection{Agentic framework for synthetic data generation}
\label{app:agentic_framework}
\input{\PaperRoot figures/fig_agentic}
\figref{fig:agentic_framework} summarizes our agentic framework for constructing outdoor training scenes.
We first use GPT-5.6~\citep{openai2026gpt56} to generate 4000 prompts describing diverse outdoor scenes from 20 categories: city streets, parks and gardens, suburban houses, landmarks, and so on. Then we use Ideogram~\citep{ideogram-4-2026} to generate a reference image. Given each image, GPT-5.6 outputs a scene plan containing object categories, approximate object dimensions, spatial relations, and ground regions.
For object reconstruction, GPT-5.6 localizes each object in the image with a bounding box. We crop each detected object and use Nano Banana~\citep{google2025nanobananapro} to complete its occluded regions, producing an image of the full object. Trellis~2 then converts the completed object image into a 3D mesh.

Following WorldClaw~\citep{guo2026worldclaw}, we reconstruct the ground separately. MoGE-3 converts the image into a top-view representation, and Nano Banana then predicts a ground-layout map over this top view. GPT-5.6 additionally predicts terrain parameters for each ground region, which we convert into a height map. We combine the layout and height map to construct the final terrain.
We then assemble the scene using a search-based layout solver following SAGE10k~\citep{xia2026sage}. For each object, the solver searches over discrete ground positions and rotations, rejects placements that introduce object overlap, and scores the candidates according to how well they satisfy the spatial relations specified in the scene plan. After placement, we run a physics simulation to place the objects onto the reconstructed terrain.

To improve scene-level consistency with the reference image, we perform one refinement round using GPT-5.6 as a VLM critic. The critic compares the reference image against a top-view rendering and four corner-view renderings of the assembled scene. It identifies missing objects and incorrect spatial relations and updates the scene plan accordingly, after which we rerun the layout solver. The final scene is then used as the final outdoor dataset.

%% file: figures/fig_app_chunk_allo.tex
\begin{figure}[htbp]
    \centering
    \includegraphics[width=0.7\linewidth]{\PaperRoot 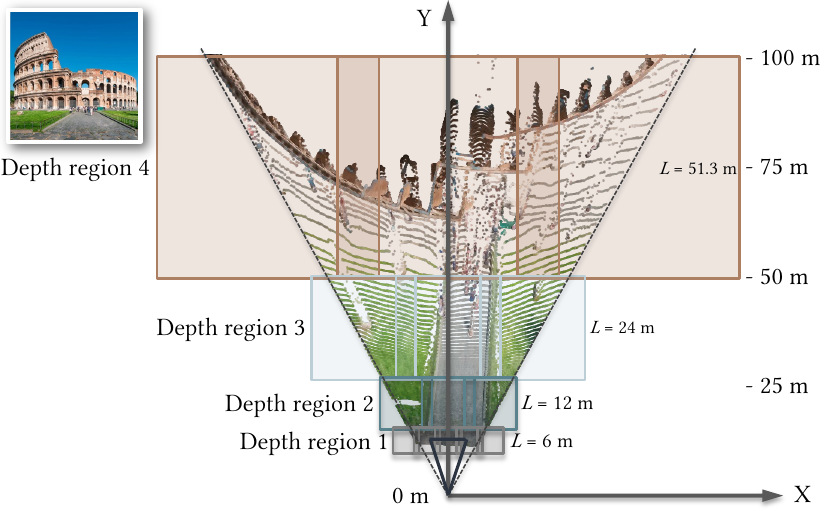}
    \caption{\textbf{Chunk allocation.}
    We partition this Colosseum into overlapping \emph{depth regions}
    and allocate chunks to cover the estimated point cloud within the
    camera frustum.}
    \label{fig:app_chunk}
\end{figure}

%% file: figures/fig_app_chunk_comparison.tex
\begin{figure}[htbp]
    \centering
    \includegraphics[width=\linewidth]{\PaperRoot 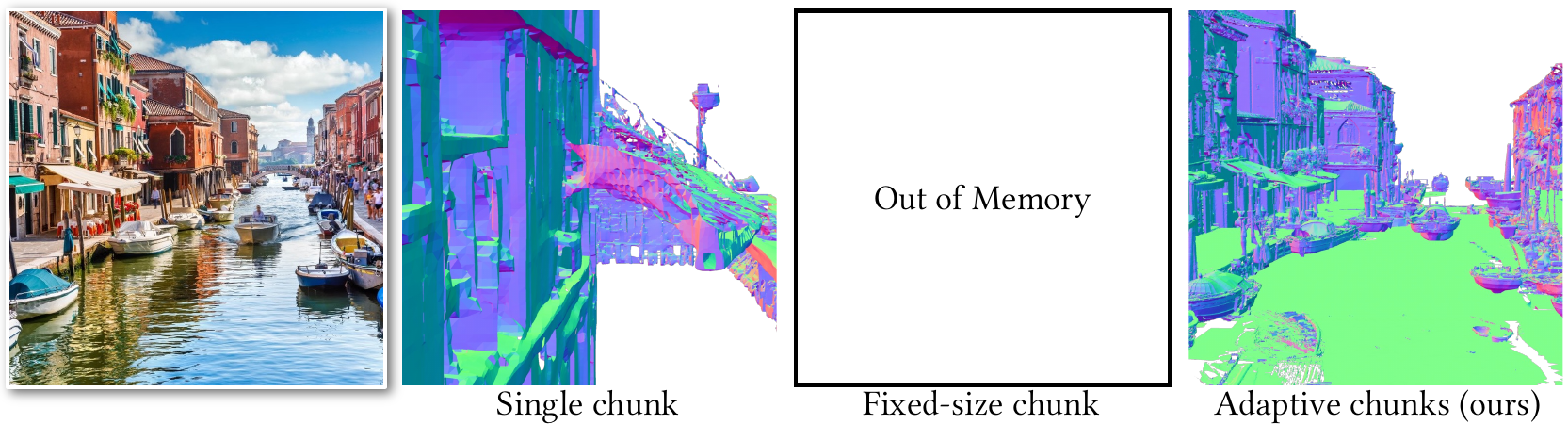}
    \caption{\textbf{Comparison of chunking strategies.}
    A single chunk compresses the entire scene into a fixed canonical
    volume, distorting depth and producing wall-like geometry.
    Fixed-size chunks require too many chunks for large scenes, and it ran out of memory on an 80G GPU. Our adaptive strategy efficiently covers
    the scene while preserving geometric detail.}
    \label{fig:app_chunk_output}
\end{figure}

%% file: figures/more_results.tex
\begin{figure*}[p]
    \centering
    \includegraphics[
        width=\textwidth,
        height=0.82\textheight,
        keepaspectratio
    ]{\PaperRoot 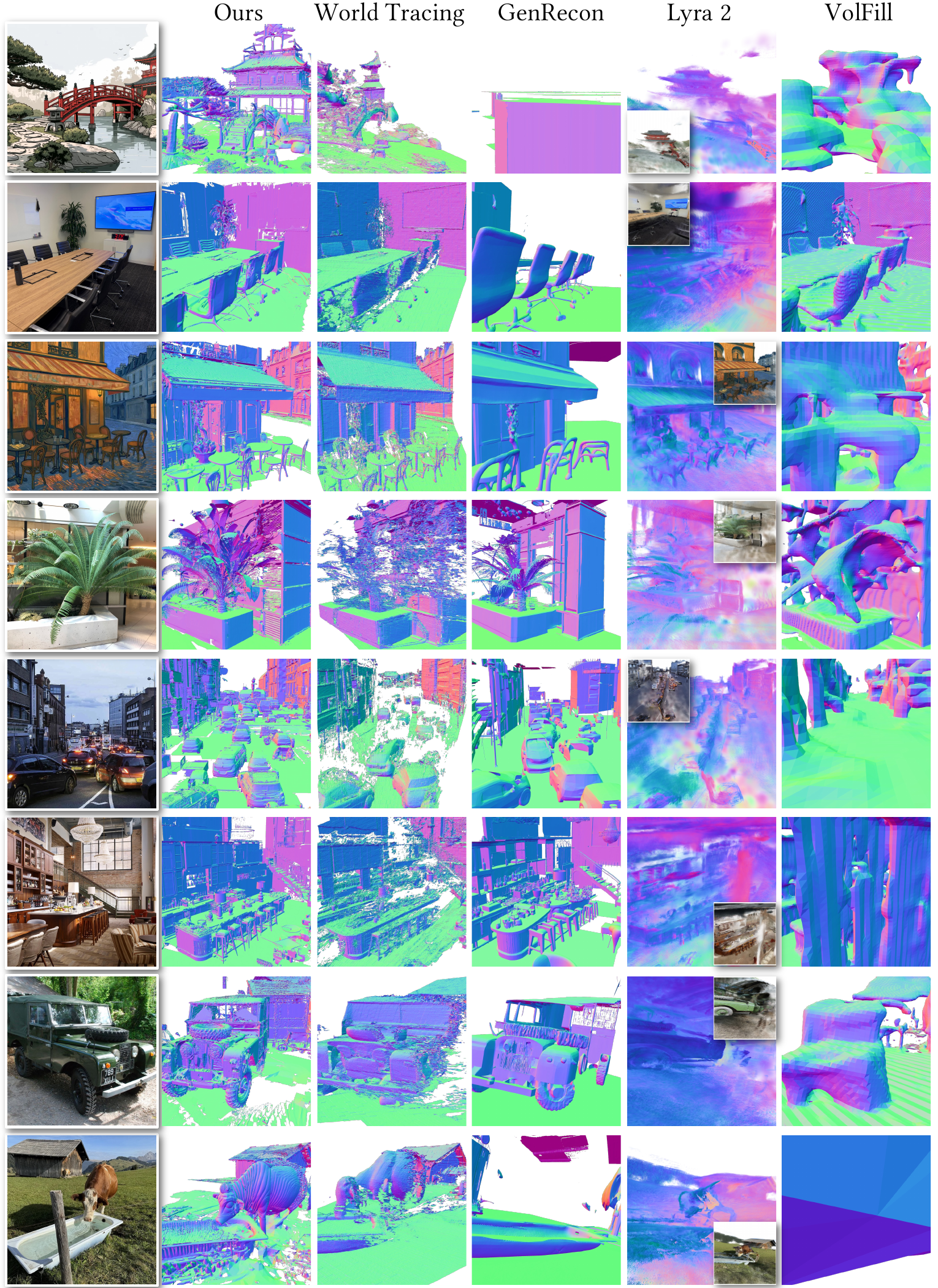}
    \caption{\textbf{Additional qualitative comparisons on diverse scenes.} We compare surface-normal renderings across indoor and outdoor environments. Our method better preserves scene layout and local geometry, and occluded regions. For Lyra 2, we overlay its RGB rendering on the normal map.}
    \label{fig:more_results}
\end{figure*}

%% file: figures/more_results2.tex
\begin{figure*}[p]
    \centering
    \includegraphics[
        width=\textwidth,
        height=0.82\textheight,
        keepaspectratio
    ]{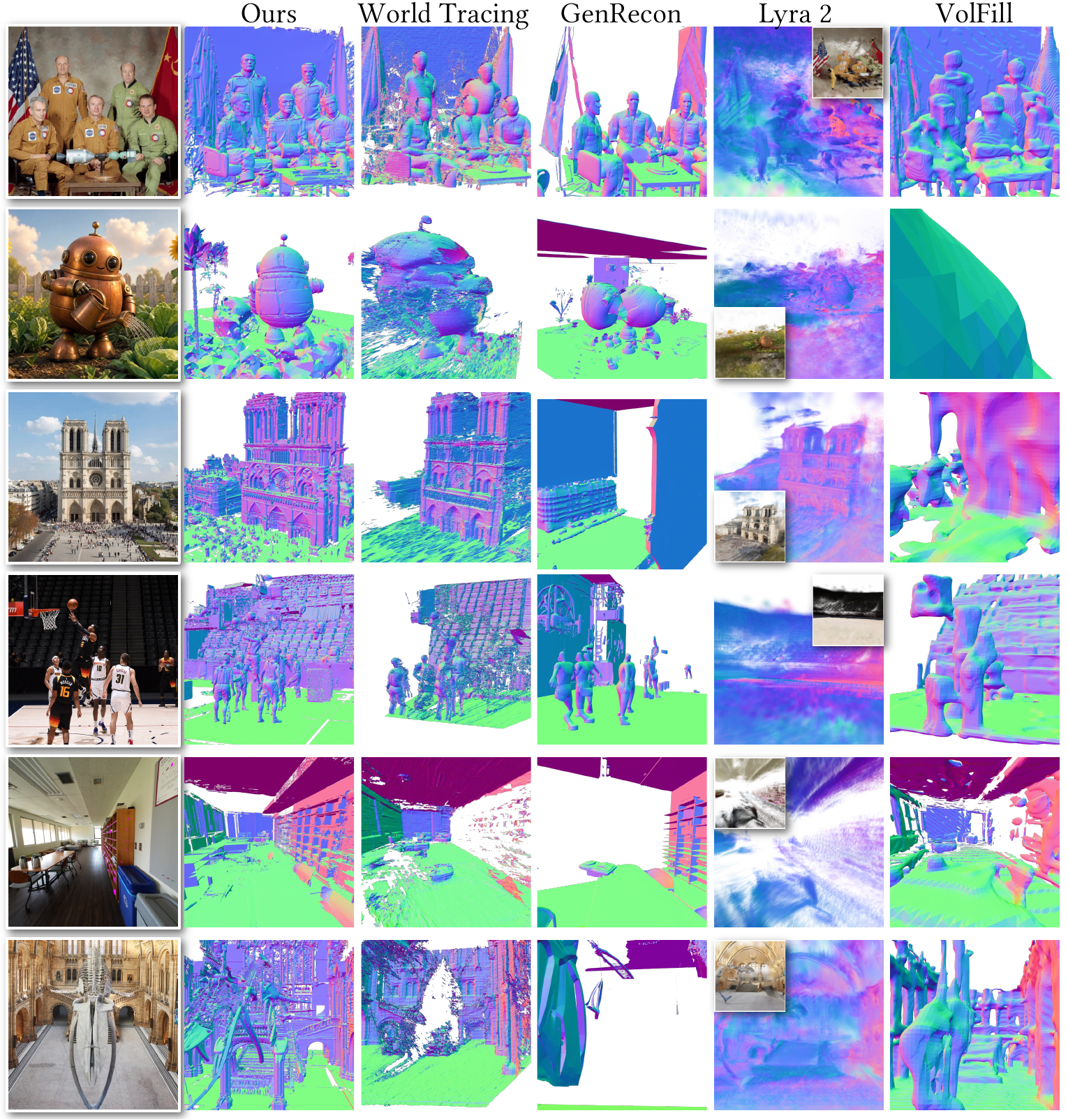}
    \caption{\textbf{Additional qualitative comparisons on diverse scenes.} We evaluate scenes containing people, animals, large architectural structures, and unusual object configurations. Our method more consistently recovers coherent geometry and scene structure, while the baselines often exhibit incomplete geometry, simplified layouts, or coarse surfaces.}
    \label{fig:more_results2}
\end{figure*}

%% file: figures/fig_agentic.tex
\begin{figure}[htbp]
\centering
    \includegraphics[width=0.98\linewidth]{\PaperRoot 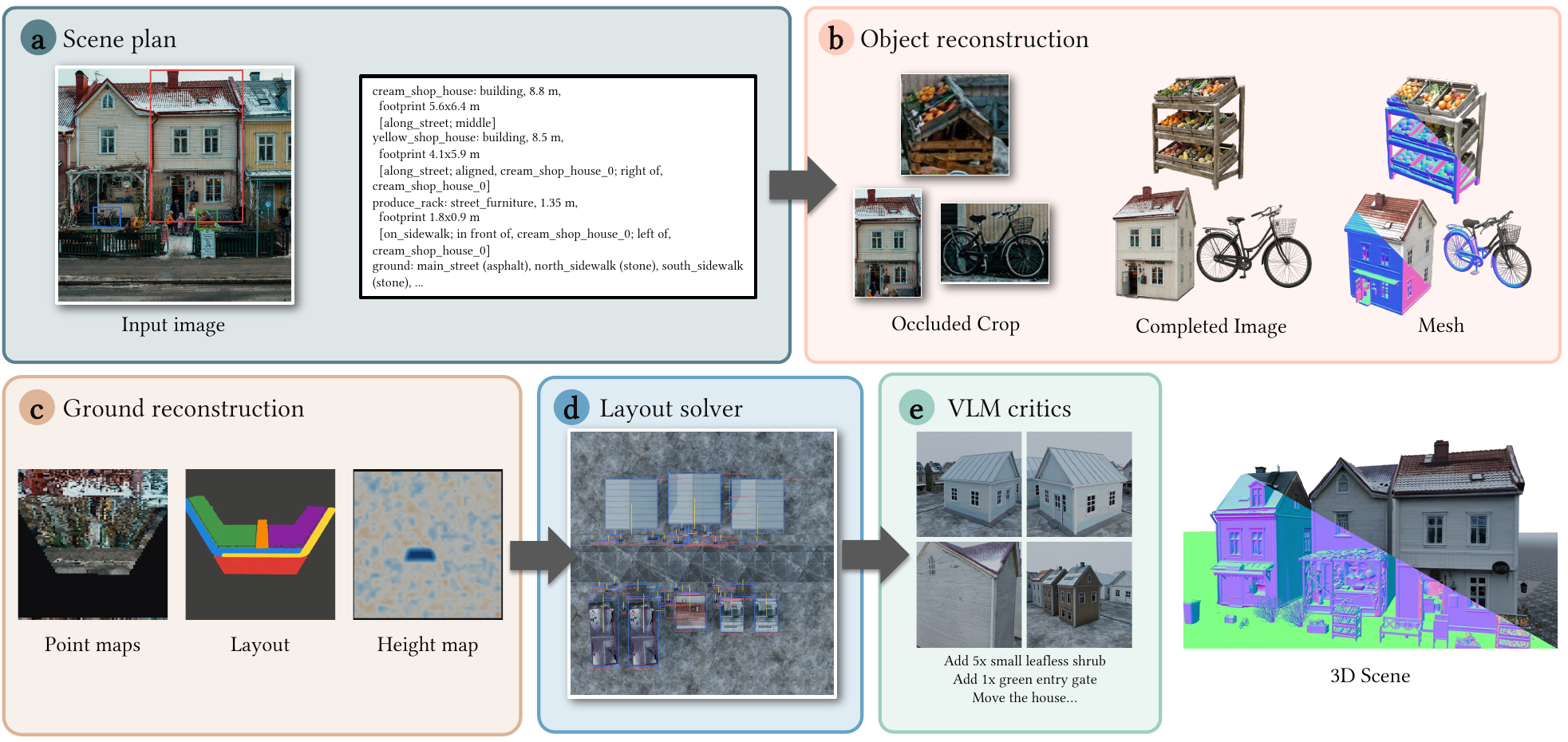}
    \caption{\textbf{Agentic framework for outdoor scene construction.}
Given an input image, (a) a VLM first produces a scene plan describing objects, their dimensions and spatial relations, and ground regions. (b) Each visible object is cropped from the image, completed to recover occluded geometry, and reconstructed into a 3D mesh. (c) Ground geometry is recovered from the estimated point map, semantic layout, and height map. (d) A layout solver places the reconstructed assets while satisfying spatial constraints and avoiding collisions. (e) A VLM critic iteratively inspects the assembled scene and proposes corrections to object placement and missing content, producing the final 3D scene.}
\label{fig:agentic_framework}
\end{figure}